\documentclass[10pt,twocolumn,letterpaper]{article}
 
\usepackage{iccv}
\usepackage{times}
\usepackage{epsfig}
\usepackage{graphicx}
\usepackage{amsmath}
\usepackage{amssymb}
\usepackage{subfig}
\usepackage{microtype}

\usepackage[pagebackref=true,breaklinks=true,letterpaper=true,colorlinks,bookmarks=false]{hyperref}

\iccvfinalcopy 

\def\iccvPaperID{2670} 

\ificcvfinal\fi

\newcommand{\localnamebig}{Convolutional Experts  Network}
\newcommand{\localnameshort}{CEN}
\newcommand{\methodnamebig}{Convolutional Experts Constrained Local Model}
\newcommand{\methodnameshort}{CE-CLM}
\newcommand{\layernamebig}{Mixture of Expert Layer}
\newcommand{\layernameshort}{ME-layer}

\newcommand{\argmin}{\operatornamewithlimits{argmin}}
\newcommand{\norm}[1]{\left\lVert#1\right\rVert}

\begin{document}

\title{\localnamebig \ for Facial Landmark Detection}

\author{Amir Zadeh$^*$\\
Carnegie Mellon University\\
5000 Forbes Ave, Pittsburgh, PA 15213, USA\\
{\tt\small abagherz@cs.cmu.edu}
\and
Tadas Baltru\v{s}aitis$^*$\\
Carnegie Mellon University\\
5000 Forbes Ave, Pittsburgh, PA 15213, USA\\
{\tt\small tbaltrus@cs.cmu.edu}
\and
Louis-Philippe Morency\\
Carnegie Mellon University\\
5000 Forbes Ave, Pittsburgh, PA 15213, USA\\
{\tt\small morency@cs.cmu.edu}
}

\maketitle


\newcommand\blfootnote[1]{%
  \begingroup
  \renewcommand\thefootnote{}\footnote{#1}%
  \addtocounter{footnote}{-1}%
  \endgroup
}
\blfootnote{$^*$ means equal contribution}
\begin{abstract}
Constrained Local Models (CLMs) are a well-established family of methods for facial landmark detection. 
However, they have recently fallen out of favor to cascaded regression-based approaches.
This is in part due to the inability of existing CLM local detectors to model the very complex individual landmark appearance that is affected by expression, illumination, facial hair, makeup, and accessories. 
In our work, we present a novel local detector -- \localnamebig \ (\localnameshort) -- that brings together the advantages of neural architectures and mixtures of experts in an end-to-end framework.
We further propose a \methodnamebig \ (\methodnameshort) algorithm that uses \localnameshort \ as local detectors. 
We demonstrate that our proposed \methodnameshort \ algorithm  outperforms competitive state-of-the-art baselines for facial landmark detection by a large margin on four publicly-available datasets. 
Our approach is especially accurate and robust on challenging profile images.


\end{abstract}
\section{Introduction}

\begin{figure}[th]
\centering
\includegraphics[width=0.96\linewidth]{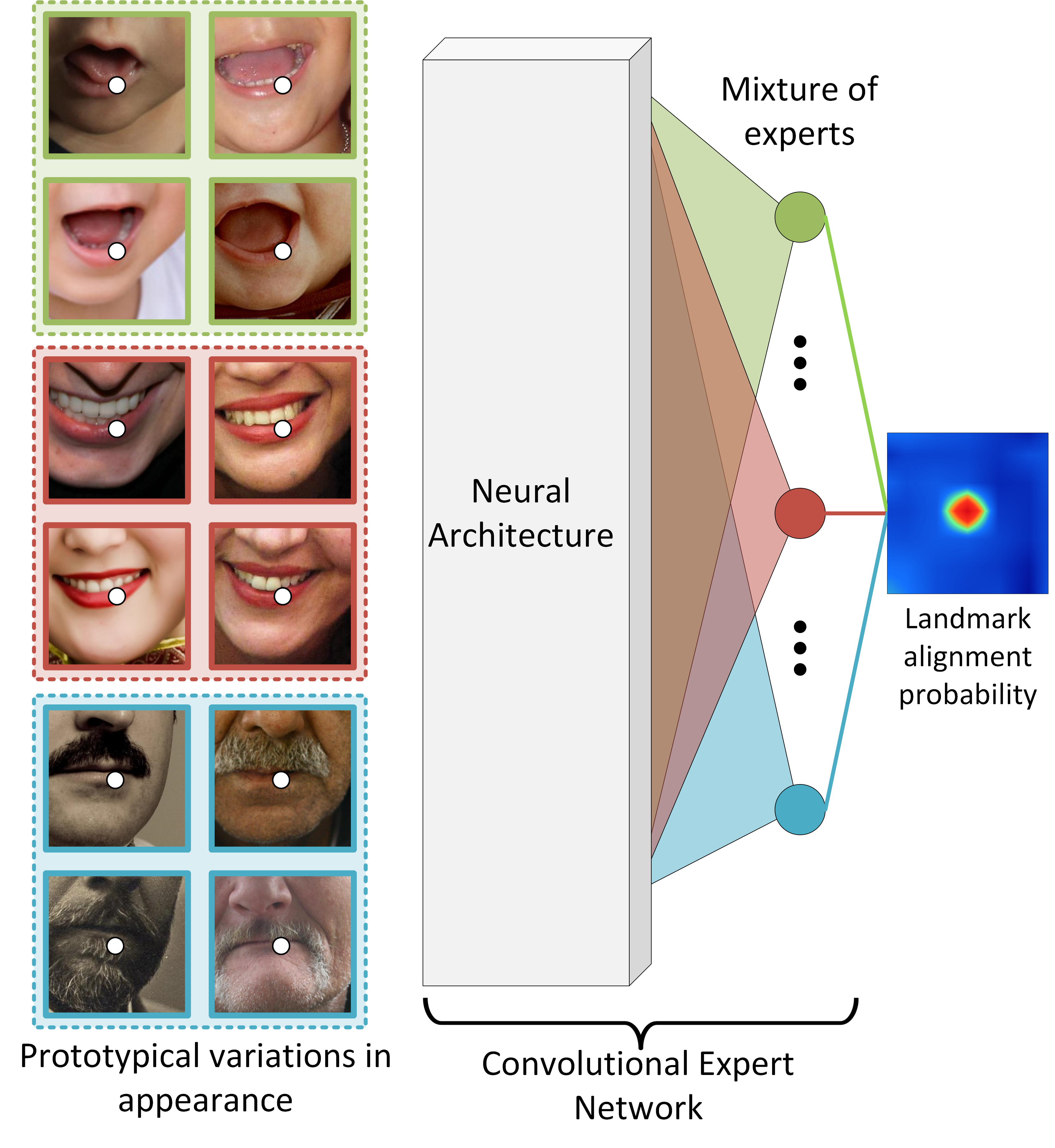}

\caption{The appearance of a facial landmark naturally clusters around a set of appearance prototypes (such as facial hair, expressions, make-up etc.). In order to model such appearance variations effectively we introduce the \localnamebig \ (\localnameshort) that brings together the advantages of neural architectures and mixtures of experts to model landmark alignment probability. \label{fig:teaser}}
\end{figure}

Facial landmark detection is an essential initial step for a number of research areas such as facial expression analysis, face 3D modeling, facial attribute analysis, multimodal sentiment analysis, emotion recognition and person identification \cite{Cristinacce2006,Martinez2016,zadeh2016mosi,Sariyanidi2014}. It is a well-researched problem with large amounts of annotated data and has seen a surge of interest in the past couple of years. 

Until recently, one of the most popular methods for facial landmark detection was the family of Constrained Local Models (CLM) \cite{Cristinacce2006,Saragih2011}. 
They model the appearance of each facial landmark individually using \emph{local} detectors and use a shape model to perform \emph{constrained} optimization.
CLMs contain many benefits and extensions that many other approaches lack: 1) modeling the appearance of each landmark individually makes CLMs robust to occlusion \cite{Asthana2013,Saragih2011}; 2) natural extension to a 3D shape model and multi-view local detectors allow CLMs to deal naturally with pose variations \cite{Saragih2011,Rajamanoharan2015} and landmark self-occlusions \cite{Baltrusaitis2014}; 3) the Expectation Maximization-based model leads to smoothness of tracking in videos \cite{Saragih2011}. 
This makes them a very appealing facial landmark detection and tracking method.



Despite these benefits, CLMs have been recently outperformed by various cascaded regression models \cite{Xiong2013,Zhu2015}.  
We believe that the relative under-performance of CLM based methods was due to the use of local detectors that are not able to model the complex variation of local landmark appearance as shown in Figure \ref{fig:teaser}. 
A robust and accurate local detector should explicitly model these different \emph{appearance prototypes} present in the same landmark. 

This paper is an extended version of a CVPR-W submission which we introduce a novel local detector called \localnamebig \ (\localnameshort) that brings together the advantages of neural architectures and mixtures of experts in an end-to-end framework \cite{ceclm17}. 
\localnameshort \ is able to learn a mixture of experts that capture different appearance prototypes without the need of explicit attribute labeling. 
To tackle facial landmark detection we present \methodnamebig \  (\methodnameshort), which is a CLM model that uses \localnameshort \ as a local detector.




We evaluate both the benefits of our \localnameshort \ local detector and  \methodnameshort \ facial landmark detection algorithm through an extensive set of experiments on four publicly-available datasets, 300-W~\cite{Sagonas2013}, 300-VW~\cite{Shen2015}, IJB-FL \cite{Kim2016}, and Menpo Challenge \cite{Zafeiriou2017}. The latter two datasets include a large portion of profile face poses with extremely challenging conditions. Furthermore, we use the latter three for cross-dataset experiments.

The structure of this paper is as follows: we discuss related work in Section \ref{sec:related}, \methodnameshort \ is introduced in Section \ref{sec:DCLM}. In Section \ref{sec:experiments} we evaluate our  \localnameshort \ local detector and compare \methodnameshort \ with other facial landmark detection approaches. We conclude the paper in Section \ref{sec:conclusion}. 
\begin{figure*}[!t]
{\includegraphics[width=\linewidth]{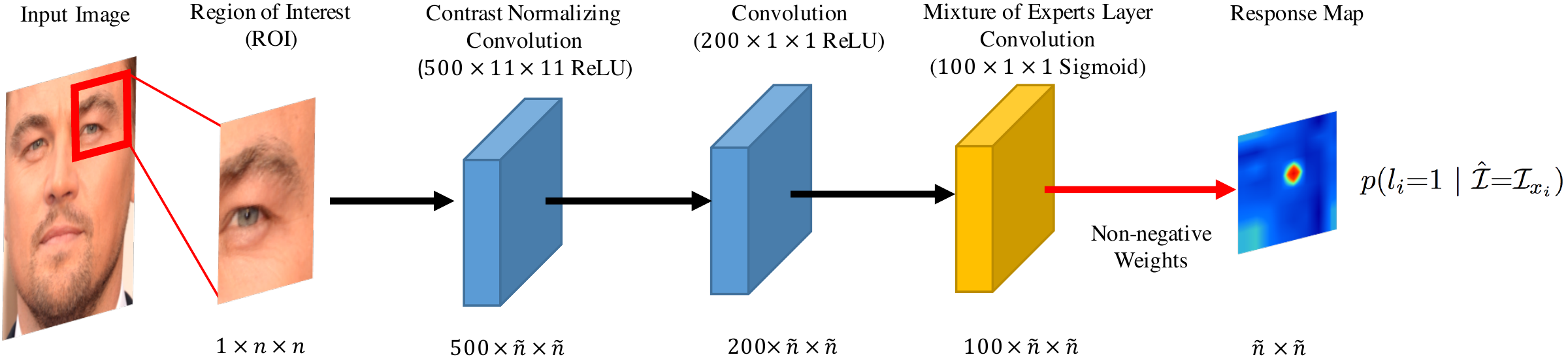}}

\caption{\label{fig:overview}Overview of our \localnamebig \ model. Input image is given and based on the estimate of the landmark position a Region of Interest with size $n \times n$ is extracted from it. This small region goes through a Contrast Normalizing Convolutional layer with kernel shape $500 \times 11 \times 11$ which performs Z-score normalization before correlation operation that outputs a $500 \times \tilde{n} \times \tilde{n}$ where $\tilde{n}=n-10$. Afterwards, the response maps are input to a convolutional layer of $200 \times 1 \times 1$ with ReLU units. \layernamebig\  (\layernameshort) learns an ensemble to capture ROI variations and uses a convolutional layer of  $100 \times 1 \times 1$ sigmoid probability decision kernels. The output response map is a non-negative and non-linear combination of neurons in \layernameshort \ using a sigmoid activation.}
\end{figure*}
\section{Related Work}
\label{sec:related}
Facial landmark detection plays a crucial role in a number of research areas and applications such as facial attribute detection \cite{Kumar2009}, facial expression analysis \cite{Martinez2016}, emotion recognition and sentiment analysis \cite{Zadeh2016,tensoremnlp17,soujanyaacl17,zadeh2015micro}, and 3D facial reconstruction \cite{Jeni2016}. 
A full review of work in facial landmark detection is outside the scope of this paper and we refer the reader to recent reviews of the field \cite{Czuprynski2014, Wang2014}.

Modern facial landmark detection approaches can be split into two major categories: \emph{model-based} and \emph{regression-based}. \emph{Model based} approaches often model both appearance and shape of facial landmarks explicitly with the latter constraining the search space and providing a form of regularization. \emph{Regression-based} approaches on the other hand do not require an explicit shape model and landmark detection is directly performed on appearance. We provide a short overview of recent model and regression based methods.

\textbf{Model-Based} approaches find the best parameters of a face model that match the appearance of an image. 
A popular model-based method is the Constrained Local Model \cite{Cristinacce2006,Saragih2011} and its various extensions such as Constrained Local Neural Fields \cite{Baltrusaitis2013} and Discriminative Response Map Fitting \cite{Asthana2013} which use more advanced methods of computing local response maps and inferring the landmark locations.  


Another noteworthy model-based approach is the mixture of trees model \cite{Zhu2012} which uses a tree based deformable parts model to jointly perform face detection, pose estimation and facial landmark detection. 
An extension of this approach is the Gauss-Newton Deformable Part Model \cite{Tzimiropoulos2014} which jointly optimizes a part-based flexible appearance model along with a global shape using Gauss-Newton optimization.
A more recently-proposed 3D Dense Face Alignment method \cite{Zhu2016} updates the parameters of a 3D Morphable Model \cite{Blanz1999} using a CNN and has shown good performance on facial landmark detection of profile faces.

%

\textbf{Regression-based} models predict the facial landmark locations directly from appearance. Majority of such approaches follow a cascaded regression framework, where the landmark detection is continually improved by applying a regressor on appearance given the current landmark estimate in explicit shape regression \cite{Cao2012}. Cascaded regression approaches include the Stochastic Descent Method (SDM) \cite{Xiong2013} which uses SIFT \cite{Lowe2004} features with linear regression to compute the shape update and Coarse-to-Fine Shape Searching (CFSS) \cite{Zhu2015} which attempts to avoid a local optima by performing a coarse to fine shape search. 
Project out Cascaded regression (PO-CR) \cite{Tzimiropoulos2015} is another cascaded regression example that updates the shape model parameters rather than predicting landmark locations directly.

Recent work has also used deep learning techniques for landmark detection. Coarse-to-Fine Auto-encoder Networks \cite{zhang2014coarse} use visual features extracted by an auto-encoder together with linear regression. Sun et al. \cite{Sun2013} proposed a CNN based cascaded regression approach for sparse landmark detection. 
Similarly, Zhang et al. \cite{Zhang2014tcdcn} proposed to use a CNN in multi-task learning framework to improve facial landmark performance by training a network to also learn facial attributes. 
Finally, Trigeorgis et al. \cite{Trigeorgis2016} proposed Mnemonic Descent Method which uses a Recurrent Neural Network to perform cascaded regression on CNN based visual features extracted around landmark locations.
\section{Convolutional Experts CLM}
\label{sec:DCLM}
\methodnamebig \ (\methodnameshort) algorithm consists of two main parts: response map computation using \localnamebig \ and shape parameter update. During the first step, individual landmark alignment is estimated independently of the position of other landmarks. During the parameter update, the positions of all landmarks are updated jointly and penalized for misaligned landmarks and irregular shapes using a point distribution model. We optimize the following objective:
\begin{equation}
	\mathbf{p}^* = \argmin_\textbf{p} \Big[\sum_{i=1}^n -\mathcal{D}_i(x_i;\mathcal{I})+\mathcal{R}(\mathbf{p})\Big]
    \label{eq:energy}
\end{equation}
above, $\mathbf{p}^*$ is the optimal set of parameters controlling the position of landmarks (see Equation \ref{eq:PDM})  with $\mathbf{p}$ being the current estimate. $\mathcal{D}_i$ is the alignment probability of landmark $i$ in location $x_i$ for input facial image $I$ (section \ref{sec:fcdp}) computed by \localnameshort. $\mathcal{R}$ is the regularization enforced by Point Distribution Model (Section \ref{sec:pdm}). The optimization of Equation \ref{eq:energy} is performed using Non-Uniform Regularized Landmark Mean Shift algorithm (Section \ref{sec:NuRLMS}). 

\subsection{\localnamebig}
\label{sec:fcdp}
The first and most important step in \methodnameshort \ algorithm is to compute a response map that helps to accurately localize individual landmarks by evaluating the landmark alignment probability at individual pixel locations. In our model this is done by \localnameshort \, which takes a $n \times n$ pixel region of interest (ROI) around the current estimate of a landmark position as input and outputs a response map evaluating landmark alignment probability at each pixel location. See Figure \ref{fig:overview} for an illustration.

In \localnameshort \ the ROI is first convolved with a contrast normalizing convolutional layer with shape $500 \times 11 \times 11$ which performs Z-score normalization before calculating correlation between input and the kernel. The output response map is then convolved with a convolutional layer of $200 \times 1 \times 1$ ReLU neurons. 

The most important layer of \localnameshort \ has the ability to model the final alignment probability through a mixture of experts that can model different landmark appearance prototypes. This is achieved by using a special neural layer called \layernamebig \  (\layernameshort) which is a convlutional layer of $100 \times 1 \times 1$ using sigmoid activation outputting individual experts vote on alignment probability (since sigmoid can be interpreted as probability). These response maps from individual experts are then combined using non-negative weights of the final layer followed by a sigmoid activation. This can be seen as a combination of experts leading to a final alignment probability. Our experiments show that \layernameshort \ is crucial for performance of the proposed \localnamebig. 

In simple terms, \localnameshort \ is given an image ROI at iteration $t$ of Equation \ref{eq:energy} as input and outputs a probabilistic response map evaluating individual landmark alignment. Thus fitting the landmark $i$ in position $x_i$ follows the equation: 
\begin{equation}
	\pi^i_{x_i}=p(l_i=1,\hat{\mathcal{I}}=\mathcal{I}_{x_i})
   \label{eq:FCDP_energy}
\end{equation}
$l_i$ is an indicator for landmark number $i$ being aligned. $\hat{\mathcal{I}}$ is the image ROI at location $x_i$ for the image $\mathcal{I}$. The response maps $\pi^i$ (of size $\tilde{n} \times  \tilde{n}$) are then used for minimizing Equation \ref{eq:energy}. The detailed network training procedure is presented in section \ref{sec:training_dpn} including chosen parameters for $n$ at train and test time. Our experiments show that making \localnameshort \ model deeper does not change the performance of the network. We study the effects of the \layernameshort \ in section \ref{sec:training_dpn} using an ablation study. 

\subsection{Point Distribution Model} 
\label{sec:pdm}
Point Distribution Models \cite{Cootes2001,Saragih2011} are used to both control the landmark locations and to regularize the shape in \methodnameshort \ framework. Irregular shapes for final detected landmarks are penalized using the term $\mathcal{R}(\mathbf{p})$ in the Equation \ref{eq:energy}. Landmark locations $\mathbf{x}_i=[x_i,y_i]^T$ are parametrized using $\mathbf{p}=[s,\mathbf{t},\mathbf{w},\mathbf{q}]$ in the following 3D PDM Equation: 
\begin{equation}
\mathbf{x}_i=s \cdot R_{2D} \cdot (\bar{\mathbf{x}}_i + \mathbf{\Phi}_i\mathbf{q}) + \mathbf{t}
\label{eq:PDM}
\end{equation}
where $\bar{\mathbf{x}}_i = [\bar{x}_i,\bar{y}_i,\bar{z}_i]^T$ is the mean value of the $i^{\mathrm{th}}$ landmark, $\mathbf{\Phi}_i$ a $3 \times m$ principal component matrix, and $\mathbf{q}$ an $m$-dimensional vector of non-rigid shape parameters; $s$, $R$ and $\mathbf{t}$ are the rigid parameters: $s$ is the scale, $R$ is a $3 \times 3$ rotation matrix defined by axis angles $\mathbf{w}=[w_x,w_y,w_z]^T$ ($R_{2D}$ are the first two rows of this matrix), and $\mathbf{t}=[t_x,t_y]^T$ is the translation.

\subsection{NU-RLMS}
\label{sec:NuRLMS}
Equation \ref{eq:energy} can be optimized using Non-Uniform Regularized Landmark Mean Shift (NU-RLMS) \cite{Baltrusaitis2013}. Given an initial \methodnameshort \ parameter estimate $\mathbf{p}$,   NU-RLMS iteratively finds an update parameter $\Delta\mathbf{p}$ such that $\mathbf{p}^*=\mathbf{p}_{0}+\Delta\mathbf{p}$, approaches the solution of Equation \ref{eq:energy}. 
NU-RLMS update finds the solution to the following problem:

\begin{equation}
	\argmin_{\Delta\mathbf{p}}\Big(\norm{\mathbf{p}_0 + \Delta\mathbf{p}}^{2}_{\Lambda^-1} + \norm {J\Delta\mathbf{p}_0 - \mathbf{v}}^2_W\Big)
    \label{eq:RLMS}
\end{equation}
where $J$ is the Jacobian of the landmark locations with respect to parameters $\mathbf{p}$. $\Lambda^{-1}$ is the matrix of priors on $\mathbf{p}$ with Gaussian prior $\mathcal{N}(\mathbf{q}; 0, \Lambda)$ for non-rigid shape and uniform for shape parameters. $W$ in Equation \ref{eq:RLMS} is a weighting matrix for weighting mean shift vectors: $W=w\cdot diag(c_1;...;c_n;c_1;...;c_n)$ and $c_i$ is the landmark detector accuracy calculated during model training based on correlation coefficient. $\mathbf{v}=[v_i]$ is the mean-shift vector calculated using a Gaussian Kernel Density Estimator using response maps of \localnameshort: 
\begin{equation}
	\mathbf{v}_i=\sum_{\mathbf{y}_i \in \Psi_i}\frac{\pi^i_{\mathbf{y}_i}\mathcal{N}(\mathbf{x_i^c; \mathbf{y}_i, \rho \mathbf{I}})}{\sum_{z_i \in \Psi_i} \pi^i_{\mathbf{z}_i}\mathcal{N}(\mathbf{x}_i^c;\mathbf{z}_i, \rho \mathbf{I})}
\end{equation}
$x_i^c$ is the current estimate for the landmark position and $\rho$ is a hyper-parameter. This leads us to the update rule of NU-RLMS:
\begin{equation}
	\Delta\mathbf{p}=-(J^TWJ+r\Lambda_{-1})(r\Lambda_{-1}\mathbf{p}-J^TW\mathbf{v})
\end{equation}

\section{Experiments}
\label{sec:experiments}
In our experiments we first evaluate the performance of \localnamebig \ and compare the performance with LNF \cite{Baltrusaitis2013} and SVR \cite{Saragih2011} local detectors (patch experts). We also evaluate the importance of the crucial \layernameshort \ for \localnameshort \ performance. 
Our final facial landmark detection experiments explore the use of our model in two settings: images and videos. All of our experiments were performed on challenging publicly available datasets and compared to a number of state-of-the-art baselines for within and cross-datasets. The \methodnameshort \ and \localnameshort\ training codes are available at 1) \url{https://github.com/A2Zadeh/CE-CLM}, 2) \url{multicomp.cs.cmu.edu/ceclm} and 3) as part of OpenFace \cite{baltruvsaitis2016openface} package \url{https://github.com/TadasBaltrusaitis/OpenFace}.

\begin{table}[t]
\centering
\caption{Comparison between \localnameshort, LNF \cite{Baltrusaitis2013} and SVR \cite{Saragih2011} using square correlation $r^2$ (higher is better) and RMSE (lower is better). To evaluate the necessity of the \layernameshort \ we also compare to \localnameshort \ (no \layernameshort), a model with no non-negative constraint on the weights of \layernameshort. Performance drop signals the crucial role of \layernameshort. }
\begin{tabular}{c|c|c}
\hline
Detector & $r^2$ & RMSE *$10^3$\\
\hline
SVR \cite{Saragih2011}& 21.31 & 66.8 \\
\hline
LNF \cite{Baltrusaitis2013}& 36.57 & 59.2 \\
\hline
\localnameshort & \textbf{64.22} & \textbf{37.9}\\
\hline
\localnameshort\ (no \layernameshort) & 23.81 & 65.11\\
\hline
\end{tabular}
\label{table:dpncompare}
\end{table}

\subsection{\localnameshort \ Experiments}
\label{sec:training_dpn}
In this section we first describe training and inference methodology of the \localnameshort \ local detector. We then compare the performance of \localnameshort \ with LNF \cite{Baltrusaitis2013} and SVR \cite{Saragih2011} patch experts followed by an ablation study to investigate the crucial role of the \layernameshort. 

\textbf{Training Procedure:} for all of the experiments \localnameshort \  was trained on LFPW and Helen training sets as well as Multi-PIE dataset. During training, if the landmark is located at the center of the $11\times 11$ convolutional region, then the probability for the landmark presence was high, otherwise low. A total of $5 \times 10^5$ convolution regions were extracted for training set and $6 \times 10^4$ were chosen for test set. We trained 28 sets of \localnameshort s \ per landmark: at seven orientations $\pm 70^{\circ}, \pm 45^{\circ}, \pm 20^{\circ}, 0$ yaw; and four scales 17, 23, 30, and 60 pixel of interocular distance. To reduce the number of local detectors that needed to be trained we mirrored the local detectors at different yaw angles and used the same expert for left and right side of the face of the frontal view. The optimizer of \localnameshort \ was Adam (\cite{adam}) with small learning rate of $5 \times 10^{-4}$ and trained for 100 epochs with mini-batches of 512 (roughly 800,000 updates per landmark). For each landmark, scale and view a \localnameshort \ local detectors has been trained. Training each \localnameshort \ model takes 6 hours on a GeForce GTX Titan X but once trained inference can be quickly done and parallelized. 
\begin{figure}[t]
\centering
\includegraphics[width=0.85\linewidth]{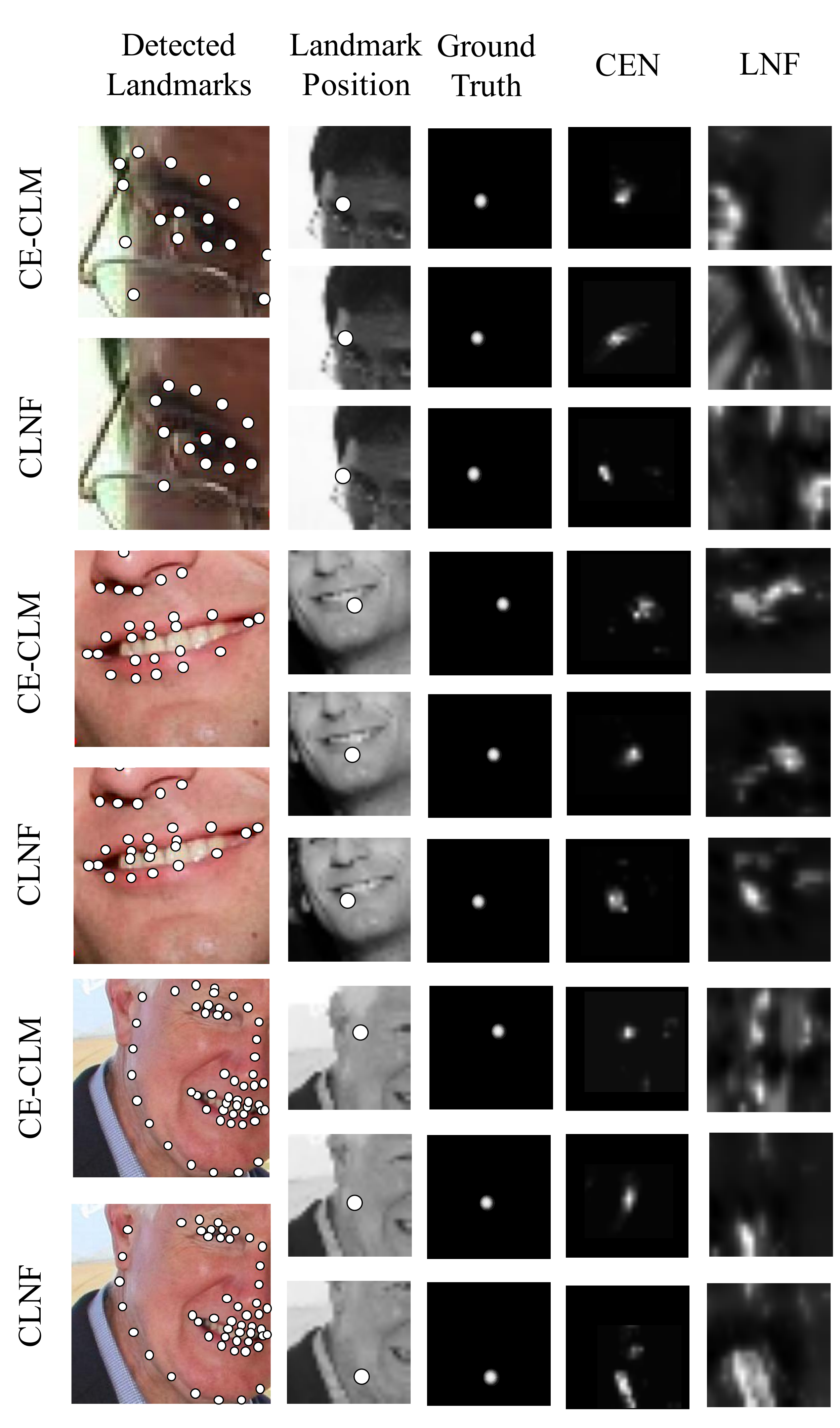}
\caption{Comparison between response maps of \localnameshort \ local detector and LNF patch experts across different landmarks. \localnameshort \ shows better localization as the landmark probability is concentrated around the correct position of the landmark. }
\label{fig:repmaps}
\end{figure}
We compare the performance improvement of \localnameshort \ local detectors over LNF and SVR patch experts. Table \ref{table:dpncompare} shows the average performance for each individual landmark. Since alignment probability inference is a regression task we use square correlation ($r ^ 2$) and RMSE between the ground truth validation set and local detector output as a measure of accuracy (higher is better for $r^2$ and lower is better for RMSE). The train and test data for all the models are the same.  On average \localnameshort \ local detector performs $75.6\%$ better than LNF and almost $200\%$ better than SVR (calculated over $r^2$), which shows a significant improvement. While this is an average, for certain landmarks, views and scales performance improvement is more than $100\%$ over LNF. This is specifically the case for $17$ pixel interocular distance scale since the \localnameshort \ is able to model the location of landmark based on a bigger appearance of landmark neighborhood in the image (more context present in the image). 

We also evaluate the importance of the \layernameshort \ in the \localnameshort \ model. Table \ref{table:dpncompare} shows the difference between \localnameshort \ and \localnameshort \ (no \layernameshort). We show that removing the non-negative constraint from the connection weights to final decision layer (essentially removing the model's capability to learn mixture of experts) and retraining the network drops the performance significantly, almost to the level of SVR. This signals that \layernameshort \ is a crucial and possibly the most important part of  \localnameshort \ model capturing ranges of variation in texture, illumination and appearance in the input support region while removing it prevents the model from dealing with these variations.

In Figure \ref{fig:repmaps} we visualize the improvement of \localnameshort \ over LNF local detectors across different landmarks such as eyebrow region, lips and face outline. The ground truth response map is a normal distribution centered around the position of landmark. The output response map from \localnameshort \ shows better certainty about the position of the landmark as its response map is more concentrated around the ground truth position. While LNF output is not showing such concentrated behavior. We therefore conclude that the major improvement from \localnameshort \ comes from accurate local detection, and this directly transfers to improvement in landmark detection task.

\subsection{\methodnameshort \ Experiments} 

In this section we first describe the datasets used to train and evaluate our \methodnameshort \  method. We then briefly discuss comparable state-of-the-art approaches for landmark detection. Finally we present the facial landmark detection results on images and videos.

\subsubsection{Datasets}

We evaluate our \methodnameshort \ on four publicly available datasets: one within-dataset evaluation (300-W), and three cross-dataset evaluations (Menpo, IJB-FL, 300-VW). We believe that the cross-dataset evaluations present the strongest case of \methodnameshort \ generalization when compared to the baselines. The datasets are described in more detail below.

\textbf{300-W} \cite{Sagonas2013,Sagonas2013a} is a meta-dataset of four different facial landmark datasets: Annotated Faces in the Wild (AFW) \cite{Zhu2012}, iBUG \cite{Tzimiropoulos2013}, and \textbf{LFPW + Helen}  \cite{Belhumeur2011,Le2012} datasets. We used the full iBUG dataset and the test partitions of LFPW and HELEN. This led to 135, 224, and 330 images for testing respectively. They all contain uncontrolled images of faces \emph{in the wild}: in indoor-outdoor environments, under varying illuminations, in presence of occlusions, under different poses, and from different quality cameras. We use the LFPW and HELEN test sets together with iBUG for model evaluation (as some baselines use AFW for training).

\textbf{Menpo} Benchmark Challenge \cite{Zafeiriou2017} dataset is a very recent comprehensive multi-pose dataset for landmark detection in images displaying arbitrary poses. The training set consists of 8979 images, of which 2300 are profile images labeled with 39 landmark points; the rest of the images are labeled with 68 landmarks. 
The images for the dataset are mainly re-annotated images of the challenging AFLW \cite{Kostinger2011} dataset.

\textbf{IJB-FL} \cite{Kim2016} is a landmark-annotated subset of IJB-A~\cite{Klare2015} -- a face recognition benchmark. It contains labels for 180 images (128 frontal and 52 profile faces). This is a challenging subset containing images in non-frontal pose, with heavy occlusion and poor picture quality. 

\textbf{300-VW}\cite{Shen2015} test set contains 64 videos labeled for 68 facial landmarks for every frame.
The test videos are categorized into three types: 1) laboratory and naturalistic well-lit conditions; 2) unconstrained conditions such as varied illumination, dark rooms and overexposed shots; 3) completely unconstrained conditions including illumination and occlusions such as occlusions by hand. 

\begin{figure}[t]
\centering
\subfloat[With face outline (68) \label{fig:300W-68}]
{\includegraphics[width=0.50\linewidth]{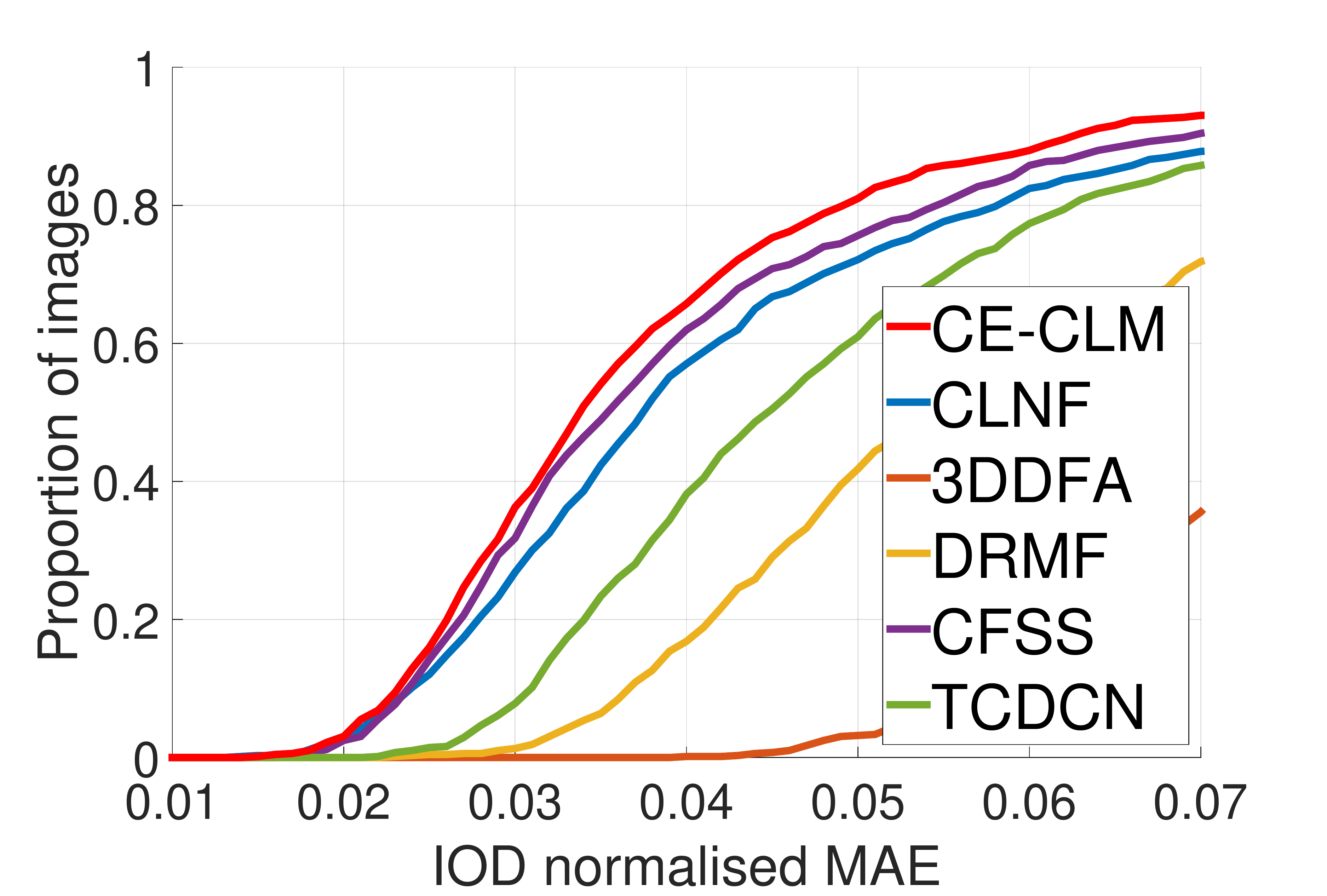}}
\subfloat[Without face outline (49) \label{fig:300W-49}]
{\includegraphics[width=0.50\linewidth]{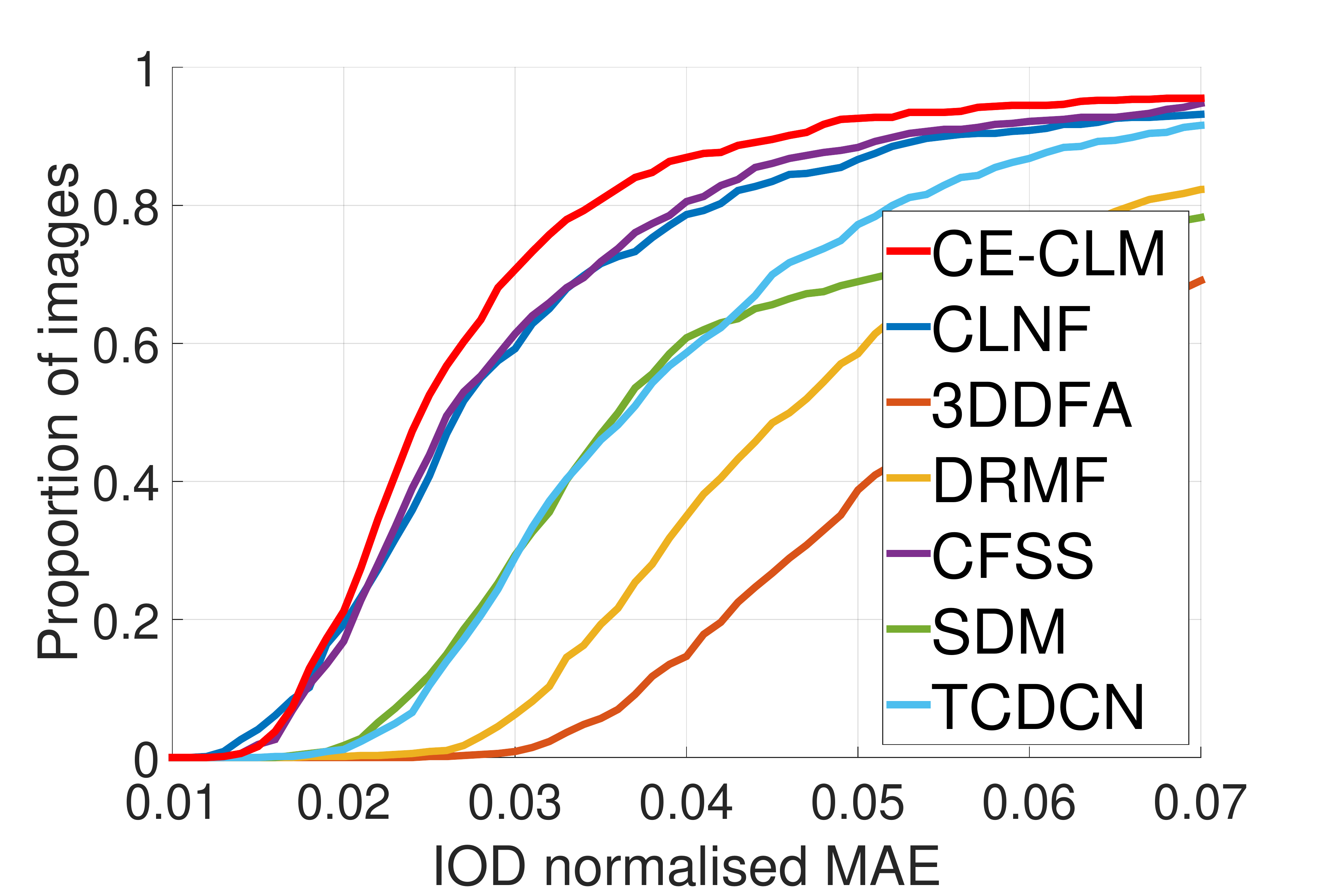}}
\caption{\label{fig:300Wres}
Cumulative error curves of IOD normalized facial landmark detection errors on the \textbf{300-W} test set -- Helen, LFPW, and iBUG. \methodnameshort \  performs better than all other approaches, especially in the difficult 68 landmark case. Best viewed in color.}
\end{figure}

\subsubsection{Baselines}

We compared our approach to a number of established baselines for the facial landmark detection task, including both cascaded regression and model based approaches. In all cases we use author provided implementations \footnote{We attempted to compare to the Mnemonic Descent Method \cite{Trigeorgis2016}, but were unable to compile the code provided by the authors due to the use of an older TensorFlow framework. As the authors do not provide results on publicly available datasets we were not able to compare our work to theirs.
}, meaning we compare to the best available version of each baseline and using the same methodology.

\textbf{CFSS} \cite{Zhu2015} -- Coarse to Fine Shape Search is a recent cascaded regression approach. It is the current state-of-the-art approach on the 300-W competition data \cite{Sagonas2013, Chrysos2016}. The model is trained on Helen and LFPW training sets and AFW.

\textbf{CLNF} is an extension of the Constrained Local Model that uses Continuous Conditional Neural Fields as patch experts \cite{Baltrusaitis2014}. 
The model was trained on LFPW and Helen training sets and CMU Multi-PIE \cite{Gross2010}.

\textbf{PO-CR} \cite{Tzimiropoulos2015} -- is a recent cascaded regression approach  that updates the shape model parameters rather than predicting landmark locations directly in a projected-out space. The model was trained on LFPW and Helen training sets.

\textbf{DRMF} -- Discriminative Response Map Fitting performs regression on patch expert response maps directly rather than using optimization over the parameter space. 
We use the implementation provided by the authors \cite{Asthana2013} that was trained on LFPW \cite{Belhumeur2011} and Multi-PIE \cite{Gross2010} datasets.

\textbf{3DDFA} -- 3D Dense Face Alignment \cite{Zhu2016} has shown state-of-the-art performance on facial landmark detection in profile images. The method uses the extended 300W-LP dataset \cite{Zhu2016} of synthesized large-pose face images from 300-W.

\textbf{CFAN} -- Coarse-to-Fine Auto-encoder Network \cite{zhang2014coarse}, uses cascaded regression on auto-encoder visual features that was trained on LFPW, HELEN and AFW.

\textbf{TCDCN} -- Tasks-Constrained Deep Convolutional Network \cite{Zhang2014tcdcn}, is another deep learning approach for facial landmark detection that uses multi-task learning to improve landmark detection performance.

\textbf{SDM} -- Supervised Descent Method is a very popular cascaded regression approach. We use implementation from the authors \cite{Xiong2013} that was trained on the Multi-PIE and LFW \cite{Huang2007} datasets.




All of the above baselines were trained to detect either landmarks \emph{without face outline} (49 or 51), or \emph{with face outline} (66 or 68). For each comparison we used the biggest set of overlapping landmarks as all the approaches share the same subset of 49 feature points.
For evaluating detections on profile images (present in IJB-FL and Menpo datasets), we use the subset of shared landmarks in ground truth images and detected ones. Since the annotations of Menpo profile faces differ slightly from the 68 landmark scheme we unify them by removing the two chin landmarks and using linear interpolation to follow the annotated curve to convert the 4 eyebrow landmarks to 5; and 10 face outline landmarks to 9. This still constitutes a fair comparison as none of the approaches (including ours) were trained on Menpo.

\begin{figure}[t]
\centering
\subfloat[Frontal faces \label{fig:menpo-front}]{\includegraphics[width=0.50\linewidth]{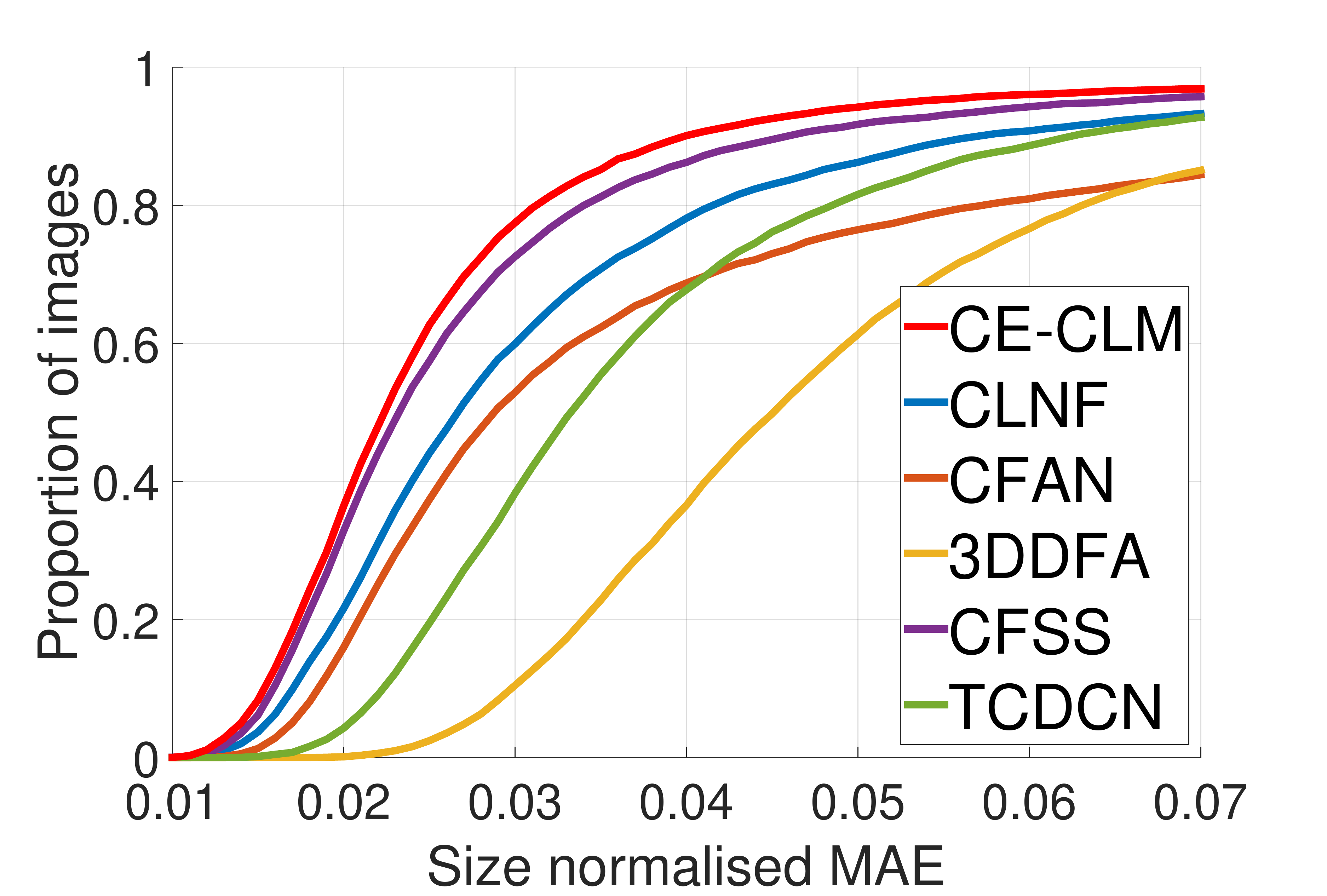}}
\subfloat[Profile faces \label{fig:menpo-prof}]{\includegraphics[width=0.50\linewidth]{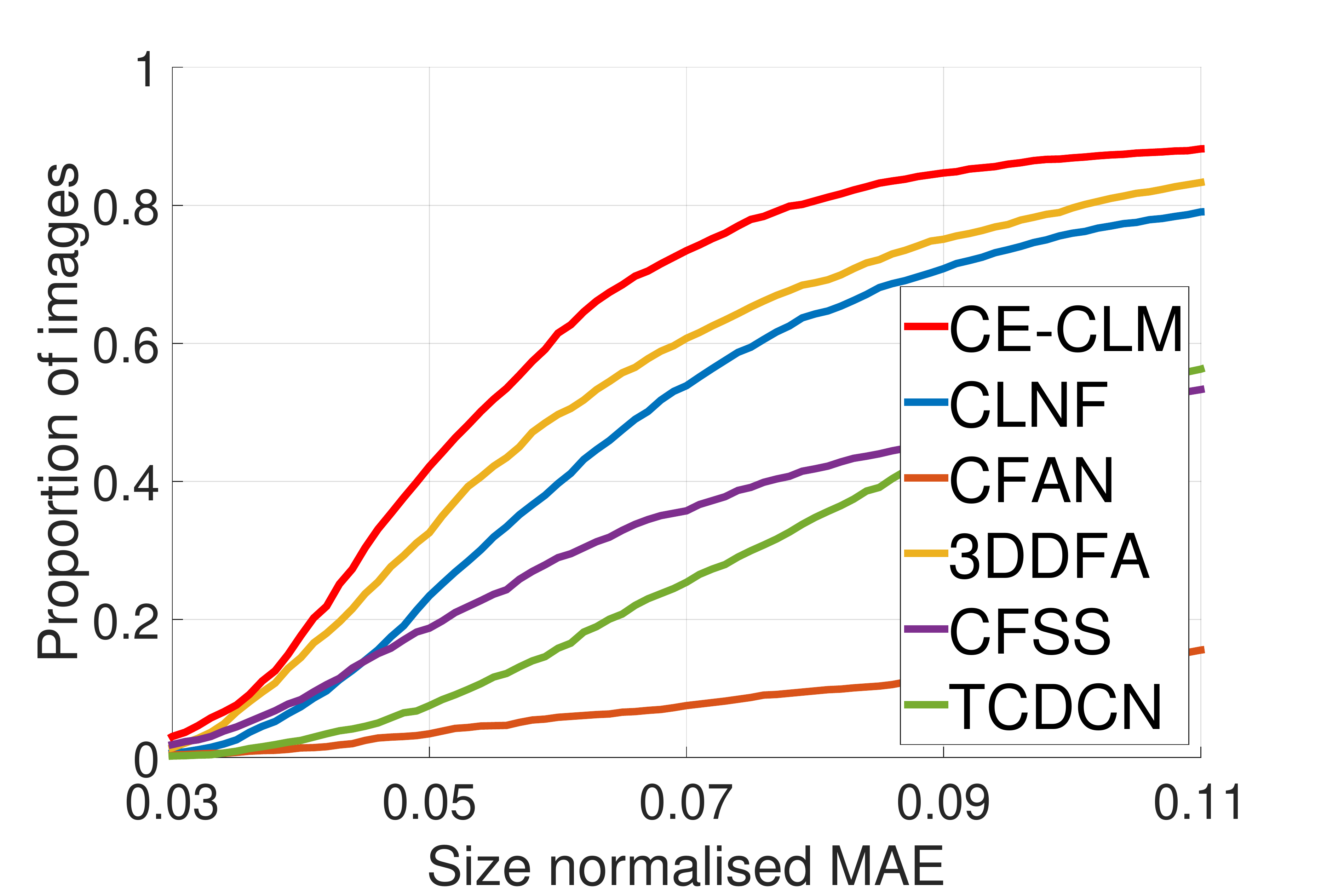}}
\caption{\label{fig:Menpo-res} Results of our facial landmark detection on the \textbf{Menpo} dataset. CE-CLM outperforms all of the baselines in both the frontal and profile image case, with a very large margin in the latter. Best viewed in color.}
\end{figure}

\begin{table}[t!]
\centering
\caption{The IOD normalized median error of landmark detection on the \textbf{300-W} dataset. We use the typical split: Comm. -- Helen and LFPW, Diff. - iBUG. \label{tab:300W}}
\begin{tabular}{c|c|c|c|c}
\hline
& \multicolumn{2}{c|}{With outline (68)} & \multicolumn{2}{c}{Without outline (49)}\\
\hline
Approach & Comm. & Diff. & Comm. & Diff. \\
\hline
CLNF \cite{Baltrusaitis2014} & 3.47 & 6.37 & 2.51 & 4.93 \\
SDM \cite{Xiong2013} & - & - & 3.31 & 10.73 \\
CFAN \cite{zhang2014coarse} & - & 8.38 & - & 6.99 \\
DRMF \cite{Asthana2013} & 4.97 & 10.36 & 4.22 & 8.64 \\
CFSS \cite{Zhu2015} & 3.20 & 5.97 & 2.46 & 4.49 \\
PO-CR \cite{Tzimiropoulos2015} & - & - & 2.67 & \textbf{3.33} \\
TCDCN \cite{Zhang2014tcdcn}  & 4.11 & 6.87 & 3.32 & 5.56 \\
3DDFA \cite{Zhu2016} & 7.27 & 12.31 & 5.17 & 8.34 \\
\hline
\methodnameshort \  & \textbf{3.14} & \textbf{5.38} & \textbf{2.30} & 3.89 
\end{tabular}
\end{table}

\begin{table}[t!]
\centering
\caption{The size normalized median landmark error on the \textbf{Menpo} dataset. We present results for profile and frontal images separately. Our approach outperforms all of the baselines in both frontal and profile images.\label{tab:menpo}}
\begin{tabular}{c|c|c|c|c}
\hline
& \multicolumn{2}{c|}{With outline (68)} & \multicolumn{2}{c}{Without outline(49)}\\
\hline
Approach & Frontal & Profile & Frontal & Profile \\
\hline
CLNF \cite{Baltrusaitis2014}& 2.66 & 6.68 & 2.10 & 4.43 \\
SDM \cite{Xiong2013} & - & - & 2.54 & 36.73 \\
CFAN \cite{zhang2014coarse} & 2.87 & 25.33  & 2.34 & 28.1 \\
DRMF \cite{Asthana2013} & - & - & 3.44 & 36.1 \\
CFSS \cite{Zhu2015}  &  2.32 & 9.99  &  1.90 & 8.42  \\
PO-CR \cite{Tzimiropoulos2015} & - & - & 2.03 & 36.0 \\
TCDCN \cite{Zhang2014tcdcn} & 3.32 & 9.82 & 2.81 & 8.69 \\
3DDFA  \cite{Zhu2016} & 4.51  & 6.02  & 3.59 & 5.47 \\
\hline
\methodnameshort \  & \textbf{2.23} & \textbf{5.39} & \textbf{1.74} & \textbf{3.32} 
\end{tabular}
\end{table}

\begin{table}[t!]
\centering
\caption{The size normalized median landmark error on the \textbf{IJB-FL} dataset. We present results for profile and frontal images separately. Our approach outperforms all of the baselines in both frontal and profile images.\label{tab:ijbfl}}
\begin{tabular}{c|c|c|c|c}
\hline
& \multicolumn{2}{c|}{With outline (68)} & \multicolumn{2}{c}{Without outline (49)}\\
\hline
Approach & Frontal & Profile & Frontal & Profile \\
\hline
CLNF \cite{Baltrusaitis2014}& 4.39 & 7.73 & 3.82 & 6.22 \\
SDM \cite{Xiong2013} & - & - & 3.93 & 30.8 \\
CFAN \cite{zhang2014coarse} & 4.89 & 20.26 & 4.37 & 22.92 \\
DRMF \cite{Asthana2013} & - & - & 4.55 & 25.52 \\
CFSS \cite{Zhu2015} & 4.16 & 7.66 & 3.57 & 6.79 \\
PO-CR \cite{Tzimiropoulos2015} & - & - & 3.73 & 21.2 \\
TCDCN\cite{Zhang2014tcdcn} & 4.91 & 9.26 & 4.53 & 9.09 \\ 
3DDFA \cite{Zhu2016} & 5.90 & 8.14 & 4.85 & 6.48 \\
\hline
\methodnameshort \  & \textbf{4.09} & \textbf{6.31} & \textbf{3.47} & \textbf{5.19} 
\end{tabular}
\end{table}

\begin{figure*}[t]
\centering
\subfloat[Category 1 \label{fig:300VW-cat1}]{\includegraphics[width=0.32\linewidth]{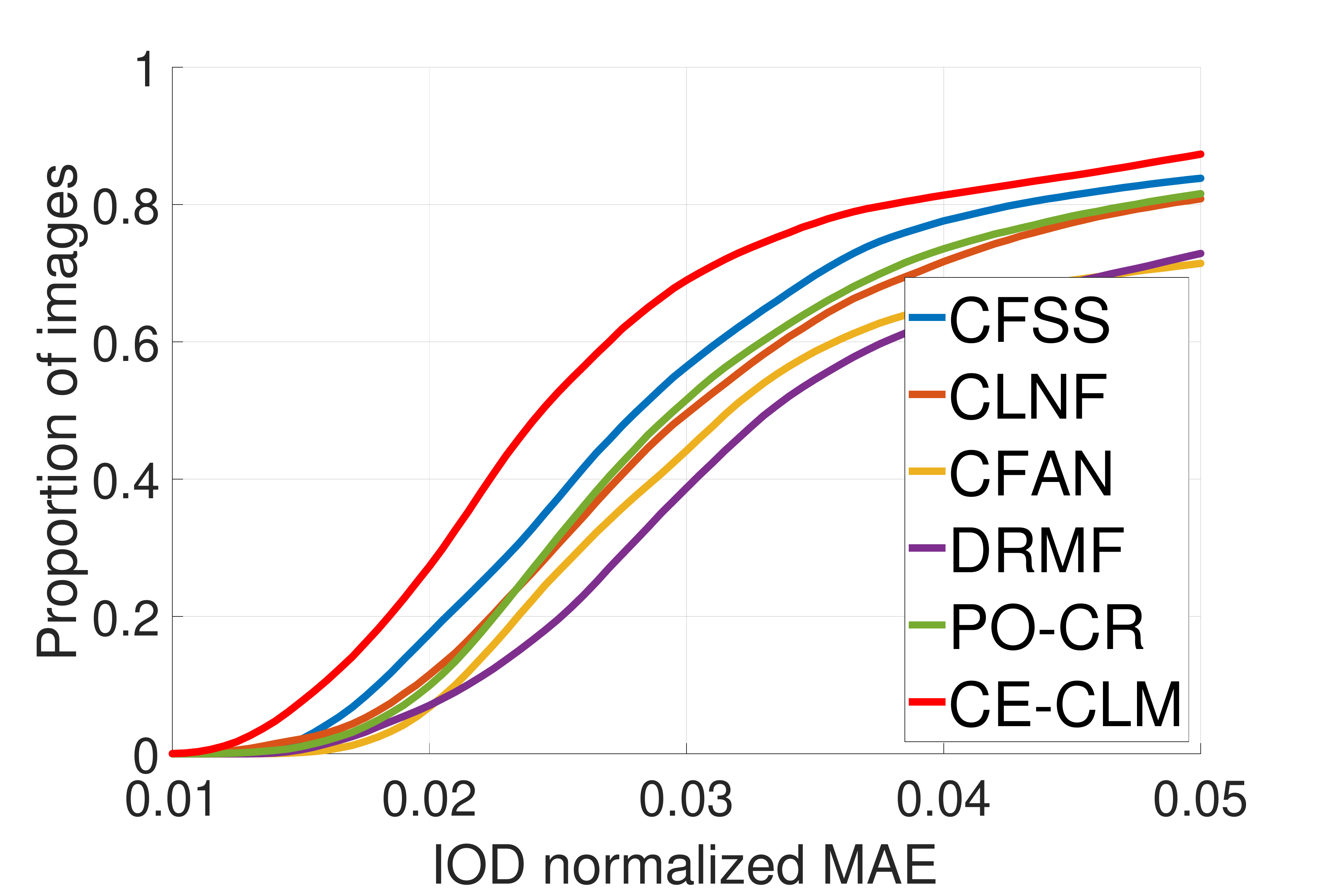}}
\subfloat[Category 2 \label{fig:300VW-cat2}]{\includegraphics[width=0.32\linewidth]{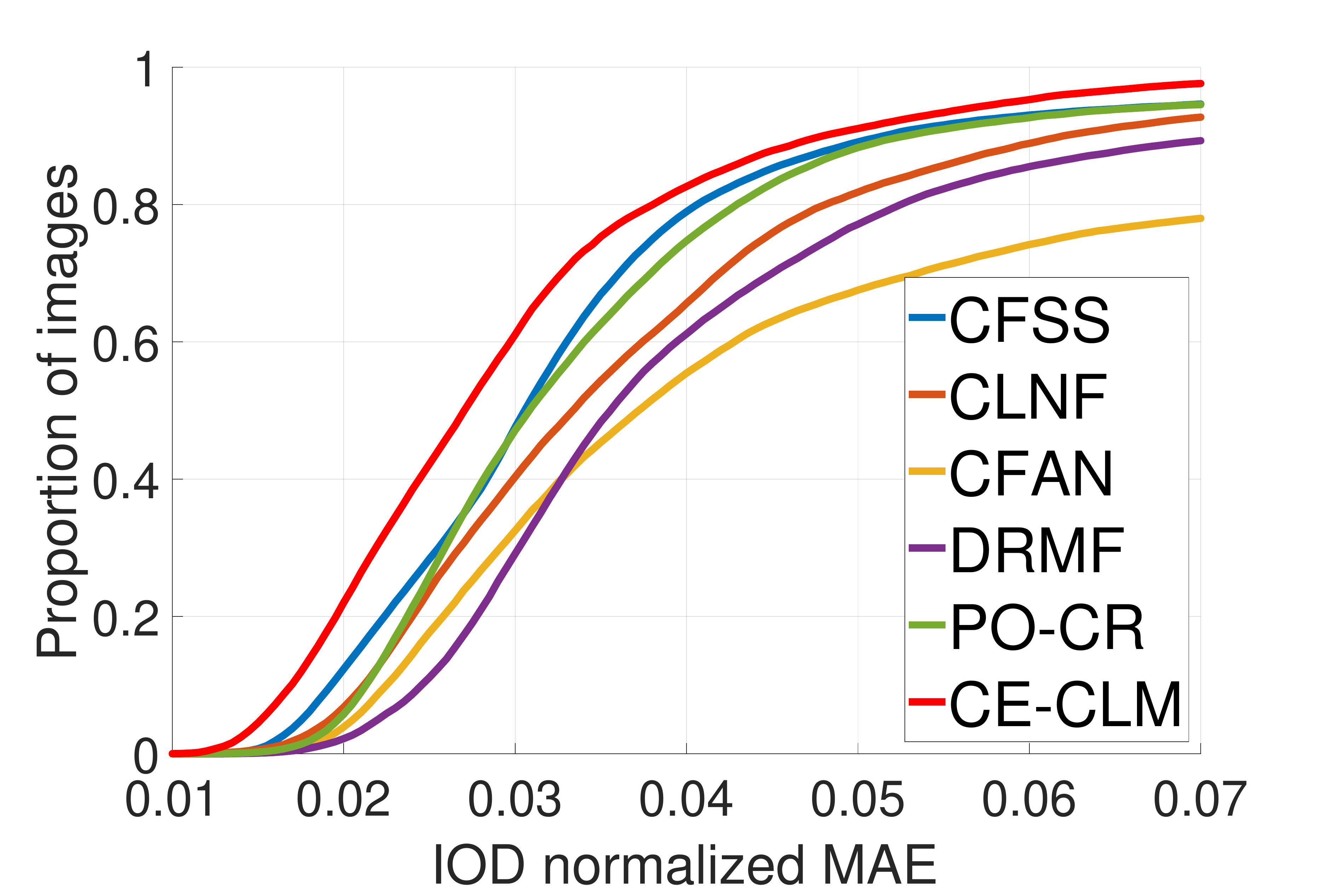}}
\subfloat[Category 3 \label{fig:300VW-cat3}]
{\includegraphics[width=0.32\linewidth]{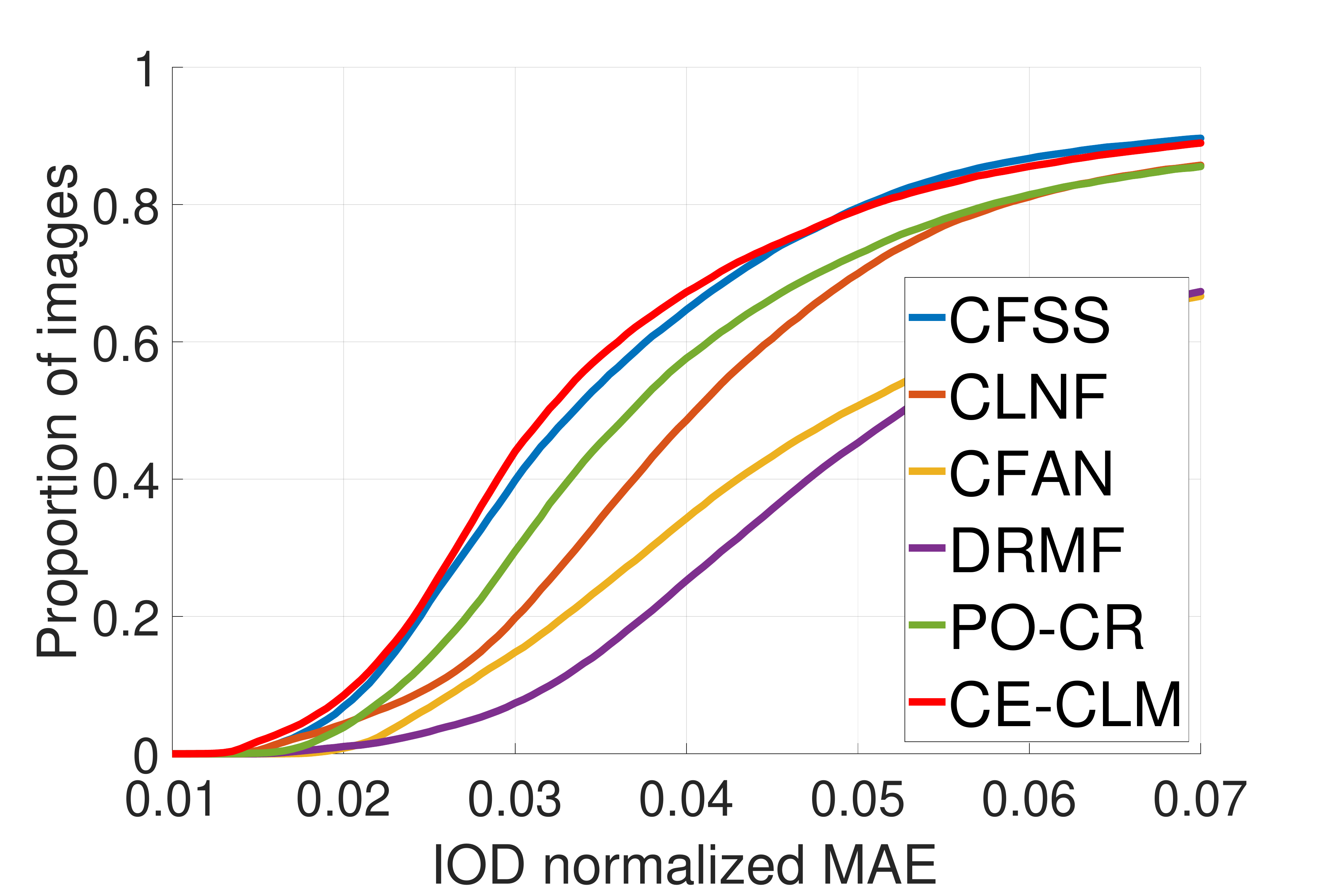}}
\caption{\label{fig:300VW-res} Results of our facial landmark detection and tracking on the \textbf{300-VW} dataset. \methodnameshort \ outperforms all of the baselines in all of the three categories. We report results on 49 inner facial landmarks. Best viewed in color.}
\end{figure*}

\subsubsection{Experimental setup}

We use the same \localnameshort \ multi-view and multi-scale local detectors as described in Section \ref{sec:training_dpn}. 
Our PDM was trained on Multi-PIE and 300-W training datasets, using non-rigid structure from motion \cite{Torresani2008}.
For model fitting we use a multi-scale approach, with a higher scale \localnameshort \ used for each iteration. For each iteration we use a progressively smaller Region of Interest -- $\{25\times 25,23\times 23, 21\times 21, 21\times 21\}$. For NU-RLMS we set $\sigma=1.85,r=32,w=2.5$ based on grid-search on the training data.
Given a bounding box, we initialized \methodnameshort \ landmark locations at seven different orientations: frontal, $\pm 30^{\circ}$ yaw, and $\pm 30^{\circ}$ pitch, and $\pm 30^{\circ}$ roll (we add four extra initializations $\pm 55^{\circ}, \pm 90^{\circ}$ yaw for Menpo and IJB-FL datasets due to large presence of profile faces). We perform early stopping and discarding of hypothesis evaluation if the converged maximum a posteriori score is above or below a threshold determined during validation. This early stopping improves the model speed by up to four times on average.
During fitting we do not compute response maps of self-occluded landmarks and do not use them for parameter update.

\begin{figure*}[t]
\includegraphics[width=\linewidth]{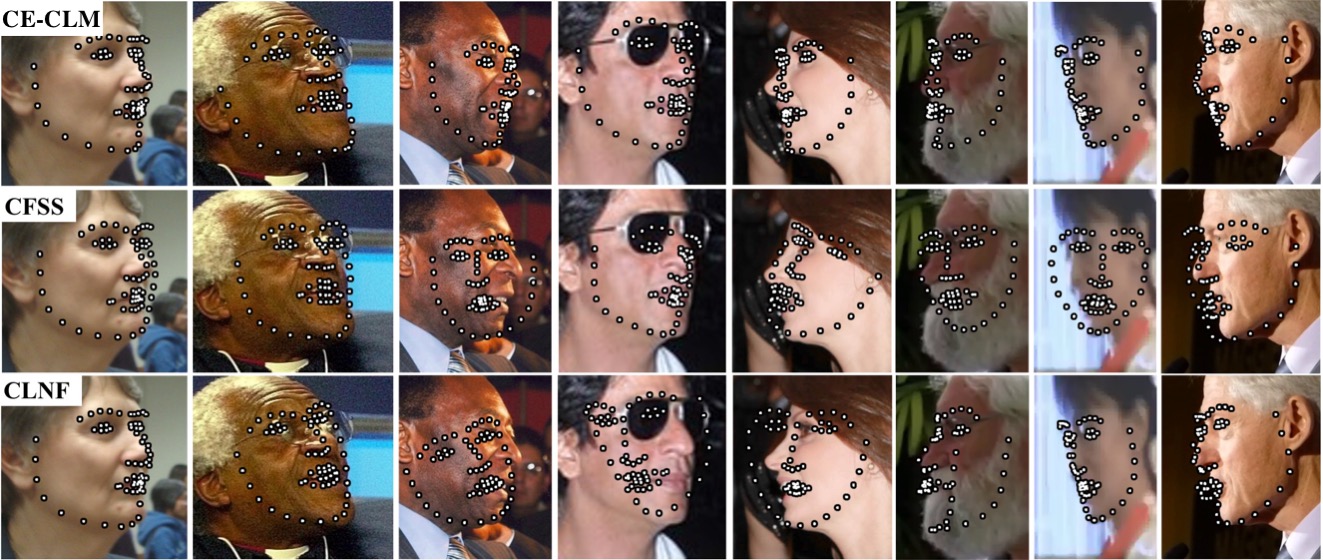}
\caption{Example images where our \methodnameshort \ approach outperforms CFSS \cite{Zhu2015} and CLNF \cite{Baltrusaitis2013}. These are challenging images due to difficulties in pose, resolution and occlusion (glasses) but \methodnameshort \ is able to align the 68 facial landmarks.}
\label{fig:comparison}
\end{figure*} 
 
For fairness of model comparison, the baselines and our model have been initialized using the same protocol. For 300-W dataset we initialized all of the approaches using the bounding boxes provided by the challenge organizers. 
For Menpo we initialized the approaches using a Multi-Task Convolutional Neural Network \cite{Zhang2016} face detector, which was able to detect faces in $96\%$ of images. We performed an affine transformation of the bounding box to match that of bounding box around the 68 facial landmarks.
For IJB-FL we initialized the approaches by generating a face bounding box by adding noise to the ground truth landmarks (based on the noise properties of the bounding boxes in 300-W dataset).
For 300-VW we detected the face in every 30th frame of each video using a Multi-Task Convolutional Neural Network \cite{Zhang2016} face detector. When the face was not detected in the frame we used the closest frame with a successful detection instead. We performed a linear mapping from the detected bounding box to a tighter fit around all 68 landmarks (as done for Menpo dataset). Each baseline was initialized from the detection and allowed to track for 30 frames, either using previously detected landmarks or using the new bounding box.

\subsubsection{Landmark Detection Results}

As common in such work we use commutative error curves of size normalized error per image to display landmark detection accuracy. We also report the size normalized median per image error. We report the median instead of the mean as the errors are not normally distributed and the mean is very susceptible to outliers.
For datasets only containing close to frontal faces (300-W and 300-VW) we normalize the error by inter-ocular distance (IOD), for images containing profile faces where one of the eyes might not be visible we instead use the average of width and height of the face.

Results of landmark detection on the \textbf{300-W} dataset can be seen in Table \ref{tab:300W} and Figure \ref{fig:300Wres}. 
Our approach outperforms all of the baselines in both the 68 and 49 point scenarios (except for PO-CR in the 49 landmark case on the iBUG dataset). 
The improved accuracy of \methodnameshort \ is especially apparent in the 68 landmark case which includes the face outline. 
This is a more difficult setup due to the ambiguity of face outline and which a lot of approaches (especially cascade regression based ones) do not tackle. 

Results of landmark detection on the \textbf{IJB-FL} dataset can be seen in Table \ref{tab:ijbfl}. 
\methodnameshort \  model outperforms all of the baselines on this difficult task as well, with a large margin for profile faces. 

Results of landmark detection on the \textbf{Menpo} dataset can be seen in Table \ref{tab:menpo} and Figure \ref{fig:Menpo-res}. 
\methodnameshort \  model outperforms all of the baselines on this difficult task as well. 
The performance improvement is especially large on profile faces, which SDM, CFAN, DRMF, and PO-CR approaches are completely unable to handle. 
We also outperform the very recent 3DDFA model which was designed for large pose face fitting.
As these results are on a cross-dataset evaluation, they demonstrate how well our method generalizes to unseen data and how well it performs on challenging profile faces (for example fits see Figure \ref{fig:comparison})

Results on landmark detection and tracking in videos on the \textbf{300-VW} dataset are displayed in Figure \ref{fig:300VW-res}.
\methodnameshort \  consistently outperforms all of the baselines in all three categories with the biggest improvement in Category 1. 
Finally, our approach outperforms the recently proposed iCCR landmark tracking method that adapts to the particular person it tracks \cite{Sanchez-Lozano2016}. However, as it is a video approach this is not a fair comparison to our work and other baselines which treat each video frame independently.
Note that our approach is consistently performing well for frontal and profile face while other approaches perform well for frontal (CFSS, PO-CR) or profile (3DDFA). This is also true across different categories of 300-VW where other approaches performance varies across categories while \methodnameshort \ consistently performs better than other approaches.


\section{Conclusion}
\label{sec:conclusion}
In this paper we introduced 
\methodnamebig \  (\methodnameshort), a new member of CLM family that uses a novel local detector called \localnamebig \ (\localnameshort). Our proposed local detector is able to deal with varying appearance of landmarks by internally learning an ensemble of detectors, thus modeling landmark appearance prototypes. This is achieved through a \layernamebig, which consists of decision neurons connected with non-negative weights to the final decision layer. In our experiments we show that this is a crucial part of \localnameshort, which outperforms previously introduced local detectors of LNF and SVR by a big margin. Due to this better performance \methodnameshort \ is able to perform better than state-of-the-art approaches on facial landmark detection and is both more accurate (Figure \ref{fig:300Wres}) and more robust, specifically in the case of profile faces  (Figure \ref{fig:Menpo-res}). Figure \ref{fig:comparison} shows a visual comparison between \methodnameshort, CFSS and CLNF landmark detection methods on a set challenging images. \methodnameshort \ is able to accurately align landmarks even in extreme profile faces.

{\small
\bibliographystyle{ieee}
\bibliography{egbib.bib}

\begin{thebibliography}{10}\itemsep=-1pt

\bibitem{Asthana2013}
A.~Asthana, S.~Zafeiriou, S.~Cheng, and M.~Pantic.
\newblock Robust discriminative response map fitting with constrained local
  models.
\newblock In {\em Proceedings of the IEEE Conference on Computer Vision and
  Pattern Recognition}, pages 3444--3451, 2013.

\bibitem{Baltrusaitis2013}
T.~Baltrusaitis, L.-P. Morency, and P.~Robinson.
\newblock Constrained local neural fields for robust facial landmark detection
  in the wild.
\newblock In {\em IEEE International Conference on Computer Vision Workshops},
  2013.

\bibitem{Baltrusaitis2014}
T.~Baltru{\v{s}}aitis, P.~Robinson, and L.-P. Morency.
\newblock Continuous conditional neural fields for structured regression.
\newblock In {\em Computer Vision--ECCV 2014}, pages 593--608. Springer, 2014.

\bibitem{baltruvsaitis2016openface}
T.~Baltru{\v{s}}aitis, P.~Robinson, and L.-P. Morency.
\newblock Openface: an open source facial behavior analysis toolkit.
\newblock In {\em Applications of Computer Vision (WACV), 2016 IEEE Winter
  Conference on}, pages 1--10. IEEE, 2016.

\bibitem{Belhumeur2011}
P.~N. Belhumeur, D.~W. Jacobs, D.~J. Kriegman, and N.~Kumar.
\newblock Localizing parts of faces using a consensus of exemplars.
\newblock In {\em CVPR}, 2011.

\bibitem{Blanz1999}
V.~Blanz and T.~Vetter.
\newblock {A Morphable Model For The Synthesis Of 3D Faces}.
\newblock In {\em SIGGRAPH}, pages 187--194, 1999.

\bibitem{Cao2012}
X.~Cao, Y.~Wei, F.~Wen, and J.~Sun.
\newblock {Face alignment by Explicit Shape Regression}.
\newblock In {\em IEEE Conference on Computer Vision and Pattern Recognition},
  pages 2887--2894. Ieee, jun 2012.

\bibitem{Chrysos2016}
G.~G. Chrysos, E.~Antonakos, P.~Snape, A.~Asthana, and S.~Zafeiriou.
\newblock {A Comprehensive Performance Evaluation of Deformable Face Tracking
  "In-the-Wild"}.
\newblock 2016.

\bibitem{Cootes2001}
T.~Cootes, G.~Edwards, and C.~Taylor.
\newblock Active appearance models.
\newblock {\em TPAMI}, 23(6):681--685, Jun 2001.

\bibitem{Cristinacce2006}
D.~Cristinacce and T.~Cootes.
\newblock Feature detection and tracking with constrained local models.
\newblock In {\em BMVC}, 2006.

\bibitem{Czuprynski2014}
B.~Czupry{\'{n}}ski and A.~Strupczewski.
\newblock {\em Active Media Technology: 10th International Conference, AMT
  2014, Warsaw, Poland, August 11-14, 2014. Proceedings}, chapter High Accuracy
  Head Pose Tracking Survey, pages 407--420.
\newblock Springer International Publishing, Cham, 2014.

\bibitem{Gross2010}
R.~Gross, I.~Matthews, J.~Cohn, T.~Kanade, and S.~Baker.
\newblock Multi-pie.
\newblock {\em IVC}, 28(5):807 -- 813, 2010.

\bibitem{Huang2007}
G.~B. Huang, M.~Ramesh, T.~Berg, and E.~Learned-Miller.
\newblock {\em Labeled Faces in the Wild: A Database for Studying Face
  Recognition in Unconstrained Environments}.
\newblock 2007.

\bibitem{Jeni2016}
L.~A. Jeni, J.~F. Cohn, and T.~Kanade.
\newblock Dense 3d face alignment from 2d video for real-time use.
\newblock {\em Image and Vision Computing}, 2016.

\bibitem{Kim2016}
K.~KangGeon, T.~Baltru\v{s}aitis, A.~Zadeh, L.-P. Morency, and G.~Medioni.
\newblock Holistically constrained local model: Going beyond frontal poses for
  facial landmark detection.
\newblock In {\em British Machine Vision Conference (BMVC)}, 2013.

\bibitem{adam}
D.~Kingma and J.~Ba.
\newblock Adam: A method for stochastic optimization
  https://arxiv.org/abs/1412.6980.

\bibitem{Klare2015}
B.~F. Klare, B.~Klein, E.~Taborsky, A.~Blanton, J.~Cheney, K.~Allen,
  P.~Grother, A.~Mah, M.~Burge, and A.~K. Jain.
\newblock Pushing the frontiers of unconstrained face detection and
  recognition: Iarpa janus benchmark a.
\newblock In {\em Computer Vision and Pattern Recognition (CVPR), 2015 IEEE
  Conference on}, pages 1931--1939. IEEE, 2015.

\bibitem{Kumar2009}
N.~Kumar, A.~C. Berg, P.~N. Belhumeur, and S.~K. Nayar.
\newblock {Attribute and simile classifiers for face verification}.
\newblock In {\em Proceedings of the IEEE International Conference on Computer
  Vision}, pages 365--372, 2009.

\bibitem{Kostinger2011}
M.~Köstinger, P.~Wohlhart, P.~M. Roth, and H.~Bischof.
\newblock Annotated facial landmarks in the wild: A large-scale, real-world
  database for facial landmark localization.
\newblock In {\em 2011 IEEE International Conference on Computer Vision
  Workshops (ICCV Workshops)}, pages 2144--2151, Nov 2011.

\bibitem{Le2012}
V.~Le, J.~Brandt, Z.~Lin, L.~Bourdev, and T.~S. Huang.
\newblock Interactive facial feature localization.
\newblock In {\em Computer Vision--ECCV 2012}, pages 679--692. Springer, 2012.

\bibitem{Lowe2004}
D.~G. Lowe.
\newblock {Distinctive image features from scale invariant keypoints}.
\newblock {\em Int'l Journal of Computer Vision}, 60:91--11020042, 2004.

\bibitem{Martinez2016}
B.~Martinez and M.~Valstar.
\newblock Advances, challenges, and opportunities in automatic facial
  expression recognition.
\newblock In B.~S. M.~Kawulok, E.~Celebi, editor, {\em Advances in Face
  Detection and Facial Image Analysis}, pages 63 -- 100. Springer, 2016.

\bibitem{soujanyaacl17}
S.~Poria, E.~Cambria, D.~Hazarika, N.~Mazumder, A.~Zadeh, and L.-P. Morency.
\newblock Context-dependent sentiment analysis in user-generated videos.
\newblock In {\em Association for Computational Linguistics}, 2017.

\bibitem{Rajamanoharan2015}
G.~Rajamanoharan and T.~F. Cootes.
\newblock Multi-view constrained local models for large head angle facial
  tracking.
\newblock In {\em The IEEE International Conference on Computer Vision (ICCV)
  Workshops}, December 2015.

\bibitem{Sagonas2013}
C.~Sagonas, G.~Tzimiropoulos, S.~Zafeiriou, and M.~Pantic.
\newblock 300 faces in-the-wild challenge: The first facial landmark
  localization challenge.
\newblock In {\em Proceedings of the IEEE International Conference on Computer
  Vision Workshops}, pages 397--403, 2013.

\bibitem{Tzimiropoulos2013}
C.~Sagonas, G.~Tzimiropoulos, S.~Zafeiriou, and M.~Pantic.
\newblock 300 faces in-the-wild challenge: The first facial landmark
  localization challenge.
\newblock In {\em ICCV}, 2013.

\bibitem{Sagonas2013a}
C.~Sagonas, G.~Tzimiropoulos, S.~Zafeiriou, and M.~Pantic.
\newblock A semi-automatic methodology for facial landmark annotation.
\newblock In {\em Proceedings of the IEE Conference on Computer Vision and
  Pattern Recognition Workshops (CVPR-W), Workshop on Analysis and Modeling of
  Faces and Gestures}, 2013.

\bibitem{Sanchez-Lozano2016}
E.~S{\'{a}}nchez-Lozano, B.~Martinez, G.~Tzimiropoulos, and M.~Valstar.
\newblock {Cascaded Continuous Regression for Real-time Incremental Face
  Tracking}.
\newblock In {\em ECCV}, 2016.

\bibitem{Saragih2011}
J.~Saragih, S.~Lucey, and J.~Cohn.
\newblock {Deformable Model Fitting by Regularized Landmark Mean-Shift}.
\newblock {\em IJCV}, 2011.

\bibitem{Sariyanidi2014}
E.~Sariyanidi, H.~Gunes, and A.~Cavallaro.
\newblock {Automatic analysis of facial affect: A survey of registration,
  representation and recognition}.
\newblock {\em IEEE TPAMI}, 2014.

\bibitem{Shen2015}
J.~Shen, S.~Zafeiriou, G.~G. Chrysos, J.~Kossaifi, G.~Tzimiropoulos, and
  M.~Pantic.
\newblock {The First Facial Landmark Tracking in-the-Wild Challenge: Benchmark
  and Results}.
\newblock {\em 2015 IEEE International Conference on Computer Vision Workshop
  (ICCVW)}, 2015.

\bibitem{Sun2013}
Y.~Sun, X.~Wang, and X.~Tang.
\newblock {Deep convolutional network cascade for facial point detection}.
\newblock {\em Proceedings of the IEEE Computer Society Conference on Computer
  Vision and Pattern Recognition}, pages 3476--3483, 2013.

\bibitem{Torresani2008}
L.~Torresani, A.~Hertzmann, and C.~Bregler.
\newblock Nonrigid structure-from-motion: Estimating shape and motion with
  hierarchical priors.
\newblock {\em TPAMI}, 30(5):878 --892, may 2008.

\bibitem{Trigeorgis2016}
G.~Trigeorgis, P.~Snape, M.~A. Nicolaou, E.~Antonakos, and S.~Zafeiriou.
\newblock {Mnemonic Descent Method: A recurrent process applied for end-to-end
  face alignment}.
\newblock In {\em CVPR}, 2016.

\bibitem{Tzimiropoulos2015}
G.~Tzimiropoulos.
\newblock {Project-Out Cascaded Regression with an application to Face
  Alignment}.
\newblock In {\em CVPR}, 2015.

\bibitem{Tzimiropoulos2014}
G.~Tzimiropoulos and M.~Pantic.
\newblock Gauss-newton deformable part models for face alignment in-the-wild.
\newblock In {\em Proceedings of the IEEE Conference on Computer Vision and
  Pattern Recognition}, pages 1851--1858, 2014.

\bibitem{Wang2014}
N.~Wang, X.~Gao, D.~Tao, and X.~Li.
\newblock {Facial Feature Point Detection: A Comprehensive Survey}.
\newblock page~32, 2014.

\bibitem{Xiong2013}
X.~Xiong and F.~Torre.
\newblock Supervised descent method and its applications to face alignment.
\newblock In {\em Proceedings of the IEEE conference on computer vision and
  pattern recognition}, pages 532--539, 2013.

\bibitem{zadeh2015micro}
A.~Zadeh.
\newblock Micro-opinion sentiment intensity analysis and summarization in
  online videos.
\newblock In {\em Proceedings of the 2015 ACM on International Conference on
  Multimodal Interaction}, pages 587--591. ACM, 2015.

\bibitem{ceclm17}
A.~Zadeh, T.~Baltru{\v{s}}aitis, and L.-P. Morency.
\newblock Convolutional experts constrained local model for facial landmark
  detection.
\newblock In {\em Computer Vision and Pattern Recognition Workshop (CVPRW)}.
  IEEE, 2017.

\bibitem{tensoremnlp17}
A.~Zadeh, M.~Chen, S.~Poria, E.~Cambria, and L.-P. Morency.
\newblock Tensor fusion network for multimodal sentiment analysis.
\newblock In {\em Empirical Methods in NLP}, 2017.

\bibitem{zadeh2016mosi}
A.~Zadeh, R.~Zellers, E.~Pincus, and L.-P. Morency.
\newblock Mosi: Multimodal corpus of sentiment intensity and subjectivity
  analysis in online opinion videos.
\newblock {\em arXiv preprint arXiv:1606.06259}, 2016.

\bibitem{Zadeh2016}
A.~Zadeh, R.~Zellers, E.~Pincus, and L.-P. Morency.
\newblock {Multimodal Sentiment Intensity Analysis in Videos: Facial Gestures
  and Verbal Messages}.
\newblock {\em 2016 IEEE Intelligent Systems}, 2016.

\bibitem{Zafeiriou2017}
S.~Zafeiriou.
\newblock The menpo facial landmark localisation challenge.
\newblock In {\em Computer Vision and Pattern Recognition Workshops (CVPRW)},
  2017.

\bibitem{zhang2014coarse}
J.~Zhang, S.~Shan, M.~Kan, and X.~Chen.
\newblock Coarse-to-fine auto-encoder networks (cfan) for real-time face
  alignment.
\newblock In {\em ECCV}. Springer, 2014.

\bibitem{Zhang2016}
K.~Zhang, Z.~Zhang, Z.~Li, and Y.~Qiao.
\newblock Joint face detection and alignment using multitask cascaded
  convolutional networks.
\newblock {\em IEEE Signal Processing Letters}, 23(10):1499--1503, Oct 2016.

\bibitem{Zhang2014tcdcn}
Z.~Zhang, P.~Luo, C.-C. Loy, and X.~Tang.
\newblock {Facial Landmark Detection by Deep Multi-task Learning}.
\newblock In {\em ECCV}, 2014.

\bibitem{Zhu2015}
S.~Zhu, C.~Li, C.~C. Loy, and X.~Tang.
\newblock {Face Alignment by Coarse-to-Fine Shape Searching}.
\newblock In {\em CVPR}, 2015.

\bibitem{Zhu2016}
X.~Zhu, Z.~Lei, X.~Liu, H.~Shi, and S.~Z. Li.
\newblock {Face Alignment Across Large Poses: A 3D Solution}.
\newblock In {\em CVPR}, 2016.

\bibitem{Zhu2012}
X.~Zhu and D.~Ramanan.
\newblock Face detection, pose estimation, and landmark localization in the
  wild.
\newblock In {\em Computer Vision and Pattern Recognition (CVPR), 2012 IEEE
  Conference on}, pages 2879--2886. IEEE, 2012.

\end{thebibliography}
}

\end{document}